%% file: AGILEIR.tex
\documentclass[conference]{IEEEtran}
\IEEEoverridecommandlockouts

\usepackage{colortbl}
\usepackage{multirow}

\usepackage{cite}
\usepackage{amsmath,amssymb,amsfonts}
\usepackage{algorithmic}
\usepackage{graphicx}
\usepackage{textcomp}
\usepackage{xcolor}
\def\BibTeX{{\rm B\kern-.05em{\sc i\kern-.025em b}\kern-.08em
    T\kern-.1667em\lower.7ex\hbox{E}\kern-.125emX}}

\usepackage{graphicx}
\usepackage{ bm, subfigure, epstopdf, url, pifont, overpic, cases}
\usepackage{latexsym, amssymb, bbding, multirow, makecell,  diagbox, enumitem, caption}
\usepackage{array, enumitem, soul}
\newcommand{\R}[1]{\textcolor[rgb]{1.00,0.00,0.00}{#1}}
\newcommand{\B}[1]{\textcolor[rgb]{0.00,0.00,1.00}{#1}}

\usepackage{arydshln}

\usepackage{multirow, multicol}
\usepackage{adjustbox}

\newlength \g

\usepackage[pagebackref=true,breaklinks=true,letterpaper=true,colorlinks,bookmarks=false]{hyperref}

\begin{document}

\title{AgileIR: Memory-Efficient Group Shifted Windows Attention for Agile Image Restoration\\
\thanks{*These authors contributed equally to this work.  \dag Corresponding author.}
}

\author{\IEEEauthorblockN{1\textsuperscript{st} Hongyi Cai* }
\IEEEauthorblockA{\textit{Faculty of Computer Science }\\
\textit{and Information Technology} \\
\textit{University of Malaya}\\
Kuala Lumpur, Malaysia \\
XCloudFance@gmail.com}
\and
\IEEEauthorblockN{2\textsuperscript{nd} Mohammad Mahdinur Rahman* }
\IEEEauthorblockA{\textit{Faculty of Computer Science }\\
\textit{and Information Technology} \\
\textit{University of Malaya}\\
Kuala Lumpur, Malaysia \\
rahmanmahdinur@gmail.com}
\and
\IEEEauthorblockN{3\textsuperscript{rd} Mohammad Shahid Akhtar \dag} 
\IEEEauthorblockA{\textit{Faculty of Computer Science }\\
\textit{and Information Technology} \\
\textit{University of Malaya}\\
Kuala Lumpur, Malaysia \\
22052133@siswa.um.edu.my}
\and
\IEEEauthorblockN{4\textsuperscript{th} Jie Li}
\IEEEauthorblockA{\textit{School of Intelligence Science}\\ \textit{and Technology} \\
\textit{University of Science and Technology Beijing}\\
Beijing, China \\
lj2085727892@163.dom}
\and
\IEEEauthorblockN{5\textsuperscript{th} Jingyu Wu}
\IEEEauthorblockA{\textit{Faculty of Art and Design} \\
\textit{Fuzhou University of International Studies }\\
\textit{and Trade}\\
Fuzhou, China \\
2303669172@qq.com}
\and
\IEEEauthorblockN{6\textsuperscript{th} Zhili Fang}
\IEEEauthorblockA{\textit{Faculty of Computer Science }\\
\textit{and Information Technology} \\
\textit{University of Malaya}\\
Kuala Lumpur, Malaysia \\
fangzhili25@gmail.com}
}

\maketitle

\begin{abstract}
Image Transformers show a magnificent success in Image Restoration tasks. Nevertheless, most of transformer-based models are strictly bounded by exorbitant memory occupancy.
Our goal is to reduce the memory consumption of Swin Transformer and at the same time speed up the model during training process. Thus, we introduce AgileIR, group shifted attention mechanism along with window attention, which sparsely simplifies the model in architecture. We propose Group Shifted Window Attention (GSWA) to decompose Shift Window Multi-head Self Attention (SW-MSA) and Window Multi-head Self Attention (W-MSA) into groups across their attention heads, contributing to shrinking memory usage in back propagation. In addition to that, we keep shifted window masking and its shifted learnable biases during training, in order to induce the model interacting across windows within the channel. We also re-allocate projection parameters to accelerate attention matrix calculation, which we found a negligible decrease in performance. As a result of experiment, compared with our baseline SwinIR and other efficient quantization models, AgileIR keeps the performance still at 32.20 dB on Set5 evaluation dataset, exceeding other methods with tailor-made efficient methods and saves over 50\% memory while a large batch size is employed.
\end{abstract}

\input{intro}

\section{Method}
\label{sec:pagestyle}

AgileIR, following the paradigm of feature extraction in SwinIR, uses Shallow Feature Extraction, Deep Feature Extraction and finally reconstruction to high-quality (HQ) images. As shown in Fig. \ref{fig:arch}, we propose a group efficient attention mechanism for shifted windows and the non-shifted windows, employing in every Agile Swin Transformer Block (ASTB).

\textbf{Feature Extraction.} Given a low-quality (LQ) image $I_{LQ} \in \mathbb{R}^ {H \times W \times 3}$, it is thereby fed into shallow feature extraction layer to extract initial overview feature of the input $F_I \in \mathbb{R} ^ {H \times W \times C}$, where C denotes channel number. The extracted feature then sequentially passes through several Agile Swin Transformer Blocks (ASTB) and patch-merging in each end. A $3 \times 3$ convolutional layer is used for shallow extraction:

\begin{align}
    F_I = Conv_{3 \times 3}(I_{LQ}).
\end{align}
Taking advantages of image transformer \cite{b1,swinfir,swinoir}, we extract deep dimensional features 
$F_D \in \mathbb{R} ^{H \times W \times C}$ from $H_{DF}$, followed by formula:
\begin{align}
    F_D = H_{DF}(F_I),
\end{align}
where $H_{DF}$ represents deep feature extraction module made by multiple ASTBs.

\textbf{Reconstruction.} After ASTB blocks processing, a pixel shuffler for up-scaling expands the feature map into 2x or more than its original input in the end for Lightweight Super Resolution tasks. We eventually reconstruct the HQ output by aggregating both deep feature and shallow features:

\begin{align}
    I_{HQ} = H_{HQR}(F_I + F_D)
\end{align}
where $I_{HQ}$ stands for the HQ output from AgileIR, $H_{HQR}$ denotes high-quality image reconstruction module.

\subsection{Agile Swin Transformer Block}
As shown in Fig. \ref{fig:arch}(a), deep feature extraction consists of multiple ASTB blocks. Akin to Swin Transformer, each Agile Swin Transformer Layer (ASTL) also integrates Group Windows-based Multi-head Self Attention (GW-MSA) and Group Shifted Window-based Multi-head Self Attention (GSW-MSA). We conclude them into Group Shifted Window Attention (GSWA) in latter discussion.

\textbf{Patch Embedding.} In ASTL, when the input image is given, AgileIR split the feature $M \times M$ non-overlapping windows and thus reshape it into a ${\frac{HW}{M^{2}}}\times M^{2}\times C$ feature, where $\frac{HW}{M^2}$ indicates the aggregated number of windows and the window size is $M \times M$. 

Afterward, embedding features pass through a multi-layer perceptron (MLP) with two fully-connected layers and GELU layer. LayerNorm layer is placed before GWSA and MLP, as well as skip connections on both modules. The following demonstrates the process:
\begin{align}
    X = GWSA(LN(X)) + X, \\
    X = MLP(LN(X)) + X.
\end{align}

\input{Benchmark1}

\subsection{Group Shifted Window Attention}
\textbf{Shifted Windows into Groups.} Attention heads in Transformer are repetitively learning similar patterns across different blocks and layers \cite{b5}. Inspired by group convolution \cite{b6}, which separates a feature map into groups and cascades them in each end, we propose Group Shifted Window Attention (GSWA) to reduce memory redundancy caused by traditional Multi-head Attention and simultaneously take full advantages of feature interaction across shifted windows in SwinIR. To commence with the module, we firstly partition the input feature into $X \in \mathbb{R} ^{{\frac{HW}{M^{2}}}\times M^{2}\times C}$, cyclically shift windows as well as applying window masks, and feed them into attention module of which split the feature into $h$ groups, as illustrated in Fig. \ref{fig:gswa}. 

The $i$-th decomposed feature in the $b$-th block is denoted as $X_{b,i}$, where $1 \leq i \leq h$. The process can be formulated as:
\begin{align}
    \tilde{X}_{b,i}& =Attn(X_{b,i}W_{b,i}^{\mathrm Q},X_{b,i}W_{b,i}^{\mathrm K},X_{b,i}W_{b,i}^{\mathrm V}), \\
    \tilde{X}_{b+1}& =Concat[\tilde{X}_{b,i}]_{i=1:h}W_{b}^{\mathrm{P}}, 
\end{align}

where $W_{i}^{\mathrm Q}$, $W_{i}^{\mathrm K}$ and $W_{i}^{\mathrm V}$ are corresponding projection layers to different subspaces of $X_{b,i}$, and $W^P_b$ sets as the final projection layers in $b$-th block, aligning the concatenated projection outputs to the same dimension as the input.

In addition, to enrich the information learned by $Q, K, V$ projection layers, each $X_{b,i}$ result is accumulated from the former subsequent head $\tilde{X}_{b,i-1}$, shown in the following:
\begin{align}
    X_{b,i}& = \tilde{X}_{b,i} + \tilde{X}_{b,{i-1}}
\end{align}

\textbf{Learnable Relative Shifted Bias.} Swin Transformer intuitively shifts bias matrices when windows shift in blocks. Learnable bias in each window increases the performance, thus we adopt $B \in \mathbb{R}^{(2M-1)\times(2M-1)}$. This relative position bias shifts along with windows shifting synchronously and therefore can be learned by GSWA.

\begin{figure}[h]
    \centering
    \includegraphics[width=0.77\linewidth]{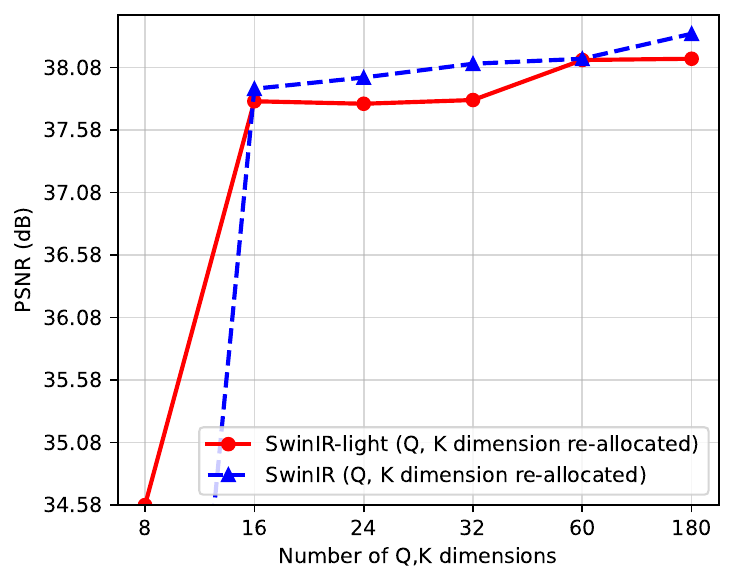}
    \caption{PSNR metric comparison of SwinIR-light and SwinIR on Set5 \cite{Set5} dataset with different Q, K dimensions. }
    \label{fig:parameter}
\end{figure}

\label{reallocation}
\textbf{Parameter Allocation.} Many studies \cite{b7} \cite{b8} have proven that the channels of $Q$ and $K$ layers are not always fully necessitated when training. We conduct several experiments to observe the performance variations by allocating them different dimension configurations. As Fig. \ref{fig:parameter} shows, traditional lightweight super resolution models mostly configure $Q$, $K$ and $V$ with 60 channels, totally counted as 180 channels. While we deduct the number of channels into 16 or 32, the model merely drops by 0.04 dB. Given these considerations, we've implemented a more efficient parameter allocation method in AgileIR to reduce memory constraints.

\section{Experiments}
\label{sec:majhead}

\input{Benchmark2}
\subsection{Settings}
\textbf{Baselines.} For a fair comparison, we select several classic Lightweight Image Restoration models (CARN \cite{ahn2018CARN}, FALSR \cite{chu2021fast}, IMDN\cite{hui2019imdn}, LAPAR\cite{li2021lapar}, SwinIR\cite{b1}) in Tab. \ref{tab:lightweight_sr_results} and efficient quantization models towards SwinIR with similar optimization regarding to memory usage, including CaDyQ \cite{cadyq}, PAMS \cite{pams} , QuantSR \cite{b2}, DoReFa \cite{dorefa} shown in Tab. \ref{tab:quant_result}. All given quantization results are either 4-bit weight with 4-bit activation (W4A4) or 8-bit weight with 8-bit activation (W8A8), since AgileIR shares similar saving of memory with them.

\textbf{Evaluation.} We evaluate reconstruction output by Peak Signal-to-Noise Ratio (PSNR) and Structural Similarity (SSIM) Index on Y channel of the YCbCr space. We conduct our experiments on Lightweight Image Super-Resolution tasks and evaluate AgileIR on Set5 \cite{Set5}, Set14 \cite{Set14}, BSD100 \cite{BSD100}, Urban100 \cite{Urban100} and Manga109 \cite{Manga109}.

\subsection{Experimental Setup}
During our training, we run AgileIR on DIV2K \cite{div2k} datasets to align with lightweight image super-resolution tasks. To gain better stability in convergence and better generalization, we adopt Charbonnier loss \cite{loss} as the loss function, along with AdamW optimizer \cite{adamw} with $\beta_1$ = 0.9 and $\beta_2$ = 0.9. The initial learning rate is 2e-4 and reduced progressively with the increment of iterations.

As for the scenario with larger resolution reconstruction, we reuse the initial weights from $\times$2 as a pre-training weight, gaining better result by learning from former tasks. The final result is obtained from weights trained for 500,000 iterations ($\times$2) and 100,000 iterations ($\times$4) with batch size 16.

In given comparisons with existent models, we divide AgileIR into two versions: AgileIR and AgileIR+. AgileIR applies 3 bottlenecks rather than 4 compared with regular SwinIR-small \cite{b1}, and yet shows a negligible performance drop. AgileIR+ applies more bottlenecks but sets the number of attention heads to 6. Both of them reduce Q and K dimension to 16 instead of being equivalent to model dimension 60, as we discussed in \ref{reallocation}. Additionally, we raise window size to 12 for better performance in GSWA. Since we apply smaller dimensions in projection layers and less complexity in terms of attention, the increase of window size is for better feature extraction and will not burden from per-window calculation.

\subsection{Results}
Selected baselines are all applying efficient techniques on SwinIR \cite{b1}. The performance of our model AgileIR+ slightly drops 0.09 $\sim$ 0.36dB ($\times$2) and 0.24 dB ($\times$4) compared with SwinIR-small, which verifies effectiveness of our method, also comparable with other SR models in Tab. \ref{tab:lightweight_sr_results}, while outperforming current state-of-arts method QuantSR-T by 0.02dB/0.0015 in Set5, 0.02dB/0.0009 in Urban100 and 0.16dB/0.0016 in Manga109. Furthermore, AgileIR also manages to exceed majority of efficient methods in each blocks. This result demonstrates superior performance with low memory usage and less computational cost.

\section{conclusion}
In this work, we explore the potential of optimizing attention heads by grouping and cascading them to reduce memory bounds and increase efficiency through architecture design. By our efforts, we manage to deduct unnecessary parameter dimensions, resulting in less runtime memory occupancy and achieve effective results on lightweight image super-resolution scenario.

\section{acknowledgement}
This research was supported in part through computational resources provided by the Data-Intensive Computing Centre, Universiti Malaya.




\vfill\pagebreak


\end{document}

%% file: intro.tex
\section{Introduction}
\label{sec:intro}

\begin{figure} [h!]
    \centering
    \includegraphics[width=0.8\linewidth]{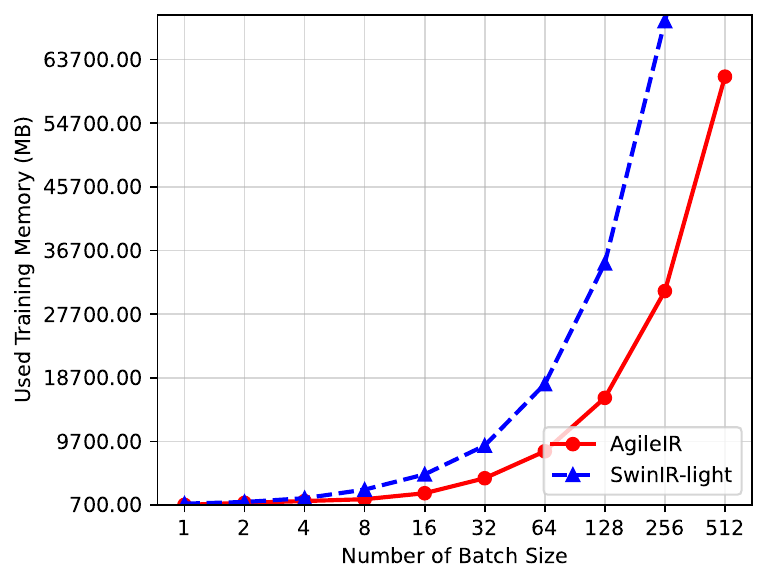}
    \caption{Shown is the comparison of memory usage in training on DIV2K \cite{div2k} between SwinIR-light \cite{b1}
    (\B{blue}) and AgileIR (\R{red}), conducted on the GPU A100 80G. Benefited from AgileIR, the training memory vastly drops 2.23X from 67.52GB to 30.23GB with the batch size set to 256. SwinIR \cite{b1} exceeds the upper bound of memory when training batch size increments to 512.}
    \label{fig:memory}
\end{figure}

CNNs and Transformer-based networks have shown good results in the image super-resolution (SR) task \cite{b3,b4,b1,restormer,uformer}. These approaches aim to generate high-resolution (HR) images from low-resolution (LR) inputs. Recent research \cite{b1} has focused on effectively combining global information with localized contextual features, resulting in more accurate HR image generation. However, while these advanced architectures and their attention mechanisms have brought significant improvements, they also come with explicit drawbacks. These models typically require large amounts of memory and have high computational complexity to achieve high-quality reconstruction. This limitation poses challenges for deploying such models on real-world edge devices. Consequently, further improvements in this area could potentially address these constraints and enhance the practical applicability of SR models.

\begin{figure*}[t]
    \centering
    \includegraphics[width=1.0\linewidth]{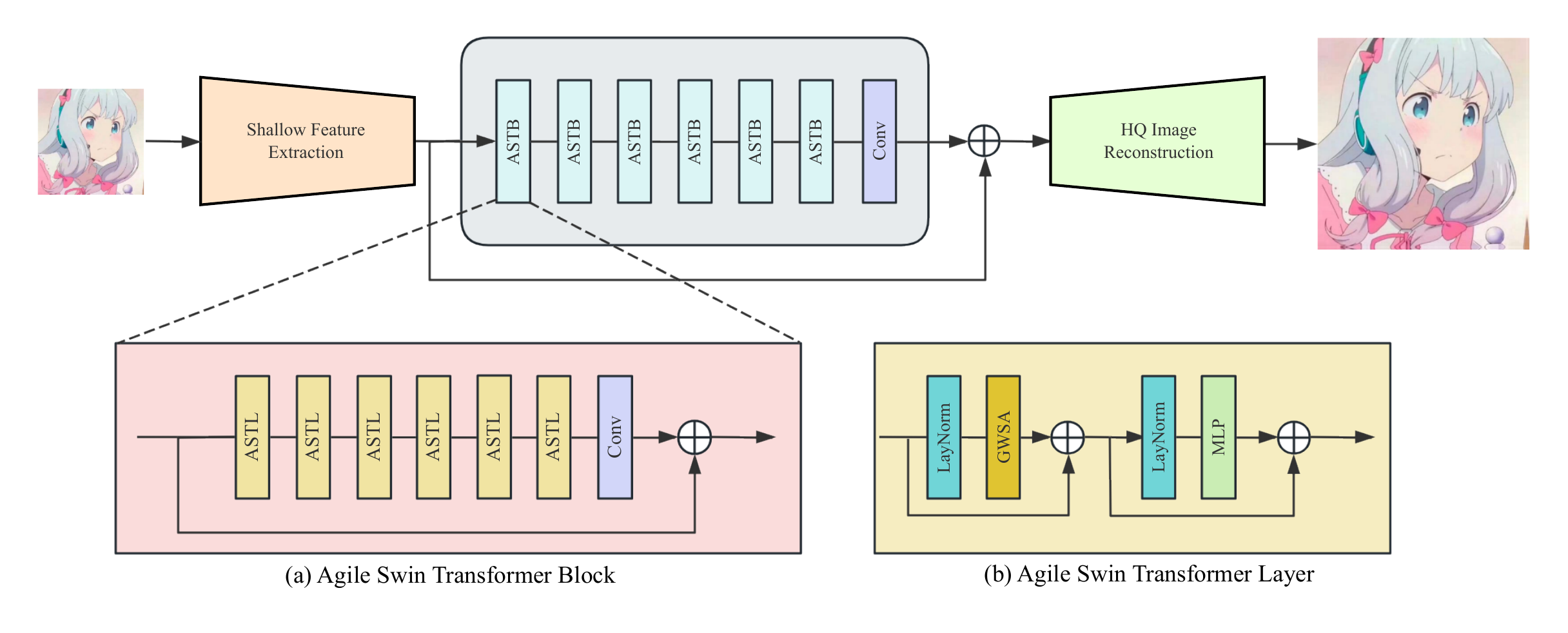}
    \caption{The overall architecture of AgileIR. ASTL represents Agile Swin Transformer Layer and HQ Image Reconstruction consists of pixel shuffler and one convolutional layer.}
    \label{fig:arch}
\end{figure*}

\begin{figure}[t]
    \centering
    \includegraphics[width=0.87\linewidth]{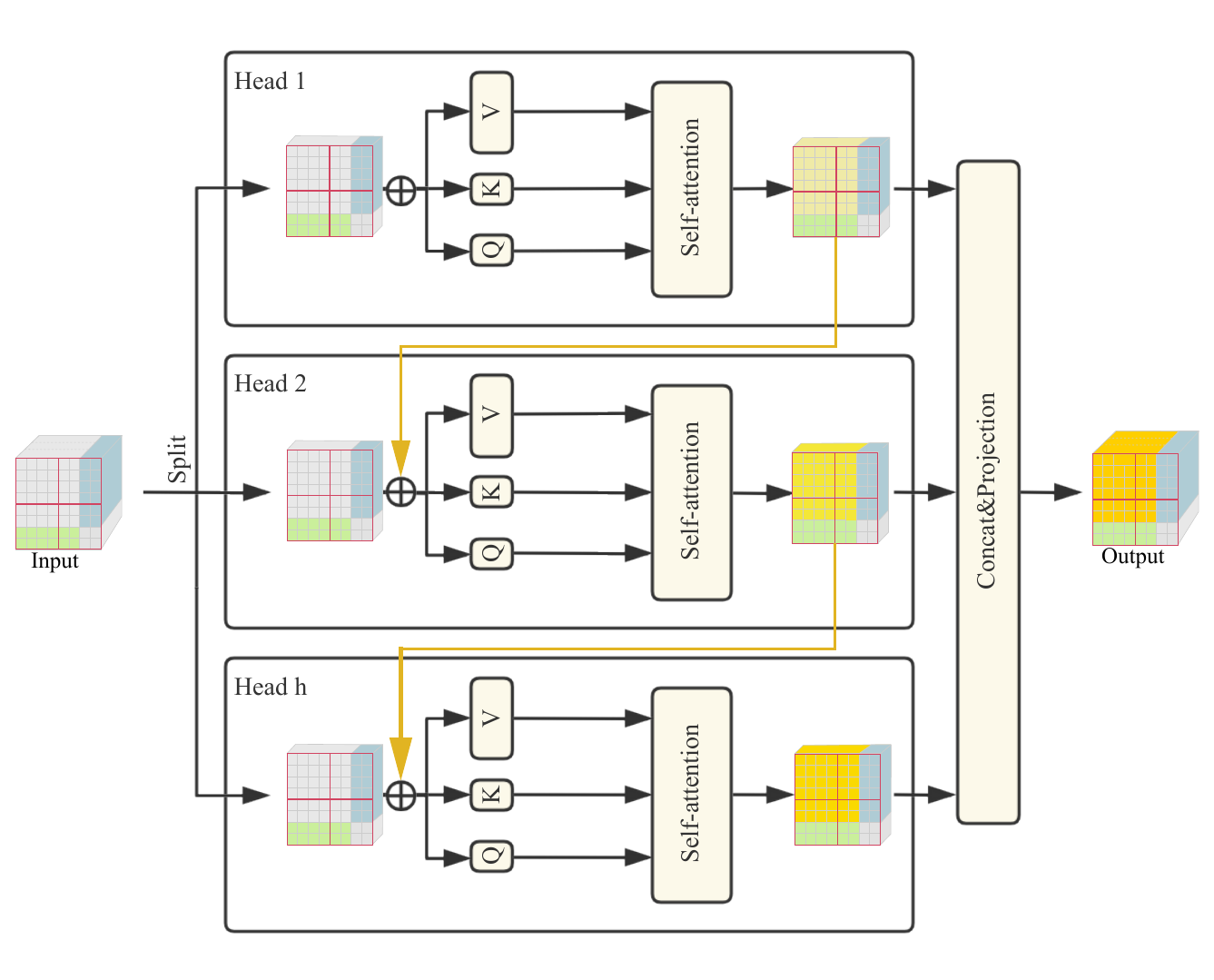}
    \caption{The Architecture of Group Shifted Window Attention.}
    \label{fig:gswa}
\end{figure}

Vision Transformers (ViT) \cite{vit} and SR models have been optimized using two main approaches: efficient architecture design \cite{b5} and model quantization techniques \cite{pams,b2,cadyq,dorefa}. These methods aim to enhance the efficiency of these models. Architecture optimization techniques focus on reducing redundant calculations and parameters, enhancing both inference and training efficiency. In contrast, model quantization directly improves speed by compressing weights and activations into lower bit representations, effectively reducing the model's overall size through clipping and mapping to a smaller bit space.

However, even though bit quantization observes the pattern of parameter distribution in ViT, it still faces performance degradation compared to their their baseline model. This mainly results from information losses caused by compressed weights during back propagation process \cite{b2}. Hence, rather than pruning in bit patterns, we present efficient group window attention and re-organize the model structure to mitigate memory-bound burdens and at the same time alleviate reconstruction quality drops from speedup strategies.

In this paper, we explore a novel cascaded layer upon the computation of swin transformer attention heads. This helps training memory of our model vastly decrease by 2X compared with original SwinIR, as shown in Fig. \ref{fig:memory}. In addition to that, we also identify redundancy in attention matrix projections via experimental results of parameter reallocation on Swin Transformer Block. 
The main contributions of our work are summarized as follows:

\begin{itemize}
    \item We propose AgileIR, extending architecture-wise sparsity based on Swin Transformer for efficient image SR regarding the trade-off between memory usage and accuracy. 
    \item We design Group Shifted Window Attention, further facilitate learnable relative biases and streamline expendable parameters to mitigate memory-bound flaws brought by Attention mechanism.
    \item We conduct experiments to compare contemporary methods of efficient super-resolution models and other lightweight SR models
\end{itemize}

%% file: Benchmark1.tex
\begin{table*}[ht!]\scriptsize
\center
\begin{center}
\caption{Quantitative comparison (average PSNR/SSIM) with other efficient W4A4 / W8A8 methods.  Best and second best performance are in \R{red} and \B{blue} colors, respectively.}
\vspace{-2mm}
\label{tab:lightweight_sr_results}

\setlength{\tabcolsep}{3.5mm}{
\begin{tabular}{|l|c|c|c|c|c|c|c|c|c|c|c|}
\hline
\multirow{2}{*}{Method} & \multirow{2}{*}{Scale} &  \multicolumn{2}{c|}{Set5~\cite{Set5}} &  \multicolumn{2}{c|}{Set14~\cite{Set14}} &  \multicolumn{2}{c|}{BSD100~\cite{BSD100}} &  \multicolumn{2}{c|}{Urban100~\cite{Urban100}} &  \multicolumn{2}{c|}{Manga109~\cite{Manga109}}  
\\
\cline{3-12}
&  & PSNR & SSIM & PSNR & SSIM & PSNR & SSIM & PSNR & SSIM & PSNR & SSIM 
\\
\hline
SwinIR (baseline) \cite{b1} & $\times$2 
& 38.14 & 0.9611 & 33.86 & 0.9206 & 32.31 & 0.9012 & 32.76 & 0.9340 & 39.12 & 0.9783
\\

\hline

DoReFa (W4A4) \cite{dorefa} & $\times$2
& 37.32
& 0.9520
& 32.90
& 0.8680
& 31.69
& 0.8504
& 30.32
& 0.8800
& 37.01
& 0.9450
\\
CADyQ (W8A8)~\cite{cadyq} & $\times$2
& 37.79
& 0.9590
& 33.37
& 0.9150
& 32.02
& 0.8980 
& 31.53
& 0.9230
& -
& -
\\
DoReFa (W8A8)~\cite{dorefa} & $\times$2
& 37.31
& 0.9510
& 32.48
& 0.9091
& 31.64
& 0.8901
& 30.18
& 0.8780
& 36.95
& 0.9440.
\\
PAMS~\cite{pams} & $\times$2
& 37.67
& 0.9588
& 33.19
& 0.9146
& 31.90
& 0.8966
& 31.10
& 0.9194
& 37.62
& 0.9400
\\
CaDyQ (W4A4) \cite{cadyq}\qquad\qquad  & $\times$2
& 37.58
& 0.9580
& 33.14
& 0.9140
& 31.87
& 0.8960
& 30.94
& 0.9170
& 37.31
& 0.9740
\\

QuantSR-T \cite{b2}\qquad\qquad  & $\times$2
& \R{38.10}
& \B{0.9604}
& \B{33.65}
& \R{0.9186}
& \R{32.21}
& \R{0.8998}
& \R{32.20}
& \R{0.9295}
& \R{38.85}
& \R{0.9774}
\\

\textbf{AgileIR} (Ours) & $\times$2
& 37.86
& 0.9600
& 33.36
& 0.9156
& 32.03
& 0.8978
& 31.54
& 0.9220
& 37.84
& 0.9755
\\
\textbf{AgileIR+} (Ours) & $\times$2
& \B{38.05}
& \R{0.9611}
& \R{33.67}
& \B{0.9176}
& \B{32.17}
& \B{0.8996}
& \B{32.13}
& \B{0.9281}
& \B{38.37}
& \R{0.9767}
\\
\hline   
\hline

SwinIR (baseline) \cite{b1} & $\times$4
& 32.44 & 0.8976 & 28.77 & 0.7858 & 27.69 & 0.7406 & 26.47 & 0.7980 & 30.92 & 0.9151
\\

\hline

DoReFa (W4A4) \cite{dorefa}& $\times$4
& 29.57
& 0.8369
& 26.82
& 0.7352
& 26.47
& 0.6971
& 23.75
& 0.6898
& 27.89
& 0.8634

\\

PAMS~\cite{pams} & $\times$4
& 31.59
& 0.8851
& 28.20
& 0.7725
& 27.32
& 0.7220
& 25.32
& 0.7624
& 28.86
& 0.8805
\\
CaDyQ (W4A4) \cite{cadyq}\qquad\qquad  & $\times$4
& 31.48
& 0.8830
& 28.05
& 0.7690
& 27.21
& 0.7240
& 25.09
& 0.7520
& 28.82
& 0.8840
\\

QuantSR-T \cite{b2}\qquad\qquad  & $\times$4
& \B{32.18}
& \B{0.8941}
& \R{28.63}
& \B{0.7822}
& \B{27.59}
& \B{0.7367}
& \B{26.11}
& \B{0.7871}
& \B{30.49}
& \B{0.9087}
\\

\textbf{AgileIR} (Ours) & $\times$4
& 31.74
& 0.8898
& 28.33
& 0.7755
& 27.40
& 0.7298
& 25.57
& 0.7668
& 29.78
& 0.8979
\\
\textbf{AgileIR+} (Ours) & $\times$4
& \R{32.20}
& \R{0.8956}
& \B{28.61}
& \R{0.7836}
& \R{27.60}
& \R{0.7376}
& \R{26.13}
& \R{0.7877}
& \R{30.65}
& \R{0.9103}
\\
\hline

\end{tabular}
}
\end{center}
\end{table*}

%% file: Benchmark2.tex
\begin{table*}[ht!]\scriptsize
\center
\begin{center}
\caption{Quantitative comparison (average PSNR/SSIM) with methods for lightweight image SR on benchmark datasets. }
\vspace{-2mm}
\label{tab:quant_result}
\begin{tabular}{|l|c|c|c|c|c|c|c|c|c|c|c|}
\hline
\multirow{2}{*}{Method} & \multirow{2}{*}{Scale} &  \multicolumn{2}{c|}{Set5~\cite{Set5}} &  \multicolumn{2}{c|}{Set14~\cite{Set14}} &  \multicolumn{2}{c|}{BSD100~\cite{BSD100}} &  \multicolumn{2}{c|}{Urban100~\cite{Urban100}} &  \multicolumn{2}{c|}{Manga109~\cite{Manga109}}  
\\
\cline{3-12}
&  & PSNR & SSIM & PSNR & SSIM & PSNR & SSIM & PSNR & SSIM & PSNR & SSIM 
\\
\hline
\hline
CARN~\cite{ahn2018CARN} & $\times$2
& 37.76
& 0.9590
& 33.52
& 0.9166
& 32.09
& 0.8978
& 31.92
& 0.9256
& 38.36
& 0.9765
\\
FALSR-A~\cite{chu2021fast} & $\times$2
& 37.82
& 0.959
& 33.55
& 0.9168
& 32.1
& 0.8987 
& 31.93 
& 0.9256
& -
& -
\\
IMDN~\cite{hui2019imdn} & $\times$2
& 38.00
& 0.9605
& 33.63
& 0.9177
& 32.19
& 0.8996
& 32.17
& 0.9283
& \B{38.88}
& \B{0.9774}
\\
LAPAR-A~\cite{li2021lapar} & $\times$2
& 38.01
& 0.9605
& 33.62
& 0.9183
& 32.19
& 0.8999
& 32.10
& 0.9283
& 38.67
& 0.9772
\\
SwinIR-small (base) \cite{b1} \qquad\qquad  & $\times$2
& \R{38.14}
& \R{0.9611}
& \R{33.86}
& \R{0.9206}
& \R{32.31}
& \R{0.9012}
& \R{32.76}
& \R{0.9340}
& \R{39.12}
& \R{0.9783}
\\

\textbf{AgileIR+} (Ours) & $\times$2
& \B{38.05}
& \R{0.9611}
& \B{33.67}
& \B{0.9176}
& 32.17
& 0.8996
& 32.13
& 0.9281
& 38.37
& 0.9767
\\
\hline   

\end{tabular}
\end{center}
\end{table*}